\documentclass[11pt]{article}
\usepackage{geometry}
\usepackage{coling2020}
\usepackage{times}
\usepackage{url}
\usepackage{latexsym}
\usepackage{microtype}
\usepackage{amssymb}
\usepackage{multirow}
\usepackage{graphicx}
\usepackage{url}
\usepackage{color,soul}
\usepackage{longtable}

\usepackage{array}

\newcolumntype{C}[1]{>{\centering\arraybackslash}m{#1}}
\newcolumntype{L}[1]{>{\left\arraybackslash}m{#1}}
\newenvironment{tight_enumerate}{
\begin{enumerate}
  \setlength{\itemsep}{0pt}
  \setlength{\parskip}{0pt}
}{\end{enumerate}}

\colingfinalcopy % Uncomment this line for all SemEval submissions

\title{SocCogCom at SemEval-2020 Task 11: Characterizing and Detecting Propaganda using Sentence-Level Emotional Salience Features}

\author{Gangeshwar Krishnamurthy, Raj Kumar Gupta, Yinping Yang \\
  Institute of High Performance Computing (IHPC), \\
  Agency for Science, Technology and Research (A*STAR), Singapore \\
  {\tt \{gangeshwark,gupta-rk,yangyp\}@ihpc.a-star.edu.sg} \\}

\date{}

\begin{document}
\maketitle
% \vspace{-15pt}
\begin{abstract}
  This paper describes a system developed for detecting propaganda techniques from news articles. We focus on examining how emotional salience features extracted from a news segment can help to characterize and predict the presence of propaganda techniques. Correlation analyses surfaced interesting patterns that, for instance, the ``loaded language" and ``slogan" techniques are negatively associated with valence and joy intensity but are positively associated with anger, fear and sadness intensity. In contrast, ``flag waving" and ``appeal to fear-prejudice" have the exact opposite pattern. Through predictive experiments, results further indicate that whereas BERT-only features obtained F1-score of 0.548, emotion intensity features and BERT hybrid features were able to obtain F1-score of 0.570, when a simple feedforward network was used as the classifier in both settings. On gold test data, our system obtained micro-averaged F1-score of 0.558 on overall detection efficacy over fourteen propaganda techniques. It performed relatively well in detecting ``loaded language" (F1 = 0.772), ``name calling and labeling" (F1 = 0.673), ``doubt" (F1 = 0.604) and ``flag waving" (F1 = 0.543).
\end{abstract}

\section{Introduction}
\label{intro}

%
% The following footnote without marker is needed for the camera-ready
% version of the paper.
% Comment out the instructions (first text) and uncomment the 8 lines
% under "final paper" for your variant of English.
% 
\blfootnote{
    %
    % for review submission
    %
    \hspace{-0.65cm}  % space normally used by the marker
    This work is licensed under a Creative Commons Attribution 4.0 International License. License details: \url{http://creativecommons.org/licenses/by/4.0/}.
    %
    % % final paper: en-uk version 
    %
    % \hspace{-0.65cm}  % space normally used by the marker
    % This work is licensed under a Creative Commons 
    % Attribution 4.0 International Licence.
    % Licence details:
    % \url{http://creativecommons.org/licenses/by/4.0/}.
    % 
    % % final paper: en-us version 
    %
    % \hspace{-0.65cm}  % space normally used by the marker
    % This work is licensed under a Creative Commons 
    % Attribution 4.0 International License.
    % License details:
    % \url{http://creativecommons.org/licenses/by/4.0/}.
}

Propaganda is studied in a wide range of social sciences disciplines, including social psychology, political science, media and mass communication, as well as advertising and marketing \cite{davison1971some,taylor2002strategic,balfour1979propaganda,mcgarry1958propaganda}. As \newcite{jowett2018propaganda} put it, propaganda is a ``deliberate, systematic attempt to shape perceptions, manipulate cognitions, and direct behavior to achieve a response that furthers the desired intent of the propagandist". To achieve the agenda, propagandists may use various influence techniques such as loaded emotive language and flag waving. Such techniques are centered on influencing the audiences' opinions and behaviors through psychological and rhetorical tricks in order to reach its purpose, such as promoting a particular politician or product in political or marketing campaigns.

The ability to automatically detect propaganda has important societal implications. For news management, propaganda detection may help publishers to quickly identify news articles that may be subjected to propagandistic characteristics that severely deviate from journalism principles. For the general public, such tools may raise awareness for social media users to stay alert of potential propagandistic content, which often may leverage non-obvious psychological tricks, and potentially mitigate the propagation of such content.

We participated in Task 11 on the detection of propaganda techniques in news articles \cite{DaSanMartinoSemeval20task11}, in particular the Technique Classification task (task TC), a multi-class classification task that aims to classify each identified text segment with the existence of a collection of fourteen propaganda techniques \cite{DaSanMartinoSemeval20task11}. Appendix A provides a summary and a distribution analysis on this task. This text segment-based ground truth data presents an advancement to this line of study with an ability to allow an algorithm to not only identify the existence of propaganda, but also to name the specific techniques. 

Our approach focuses on exploring the value of sentence-level emotional salience features to characterize propaganda techniques. From the definitions of the fourteen techniques used in \newcite{EMNLP19DaSanMartino}'s original paper, at least six techniques conceptually involve emotion-associated properties, including strong emotional connotations or emotional appeal. Consider the following examples in \newcite{EMNLP19DaSanMartino}: 
\vspace{-5pt}
\begin{tight_enumerate}
    \item ``stop those refugees; they are terrorists" [``appeal to fear-prejudice"]
    \item ``the best of the best" [``exaggeration,minimisation"]
    \item ``Entering this war will make us have a better future in our country" [``flag waving"]
    \item ``a lone lawmaker's childish shouting" [``loaded language"]
    \item ``Republican congressweasels" [``name calling,labeling"]
    \item ``Make America great again!" [``slogans"]
\end{tight_enumerate}
\vspace{-5pt}

To extract the sentence-level emotional salience features in the news segments, we leveraged \newcite{gupta2018crystalfeel}'s work which trains a collection of SVM-based algorithms, named as CrystalFeel\footnote{CrystalFeel is available at: \url{http://www.crystalfeel.socialanalyticsplus.net/}}, which detects the intensities of five emotion dimensions present in a given text message, including the sentiment valence, joy, anger, fear and sadness \cite{gupta2018crystalfeel}. As the key purpose of propaganda is to influence or persuade the audiences, our main design hypothesis is that sentence-level emotional salience features will help to characterize a few most commonly used propaganda techniques that involve a degree of emotional connotations in their language manifestations. Table \ref{table:one} illustrates the emotion intensity scores derived on six propaganda examples used in \cite{EMNLP19DaSanMartino}.

\begin{table}[h!]
\centering

\begin{tabular}{| p{0.3\linewidth} | C{0.1\linewidth} | C{0.1\linewidth}  | C{0.1\linewidth}  | C{0.1\linewidth}  | C{0.1\linewidth}  |}
% \vspace{5pt}
 \hline
 \multirow{2}{*}{\textbf{Text segment example}}
      & \multicolumn{5}{c|}{\textbf{Detected Emotion Intensity Scores (Gupta and Yang, 2018)}}\\  
      \cline{2-6}
%  \hline

  & Valence Intensity & Joy Intensity & Anger Intensity & Fear Intensity & Sadness Intensity \\  
  \hline
 ``stop those refugees; they are terrorists" [``appeal to fear-prejudice"] & 0.305 & 0.123 & \textbf{0.622} & \textbf{0.551} & \textbf{0.483} \\ 
 \hline
 ``the best of the best" [``exaggeration,minimisation"] & 0.653 & \textbf{0.520} & 0.208 & 0.183 & 0.267 \\ 
 \hline
 ``Entering this war will make us have a better future in our country" [``flag waving"]  & 0.563 & 0.344 & 0.332 & \textbf{0.406} & 0.374  \\ 
 \hline
 ``a lone lawmaker's childish shouting" [``loaded language"] & 0.323 & 0.126 & \textbf{0.516} & 0.487 &\textbf{ 0.520} \\ 
 \hline
 ``Republican congressweasels" [``name calling,labeling"]  &  0.456 & 0.216 & 0.367 & 0.371 & \textbf{0.418} \\ 
 \hline
 ``Make America great again!" [``slogans"] & 0.672 & \textbf{0.592} & 0.264 & 0.201 & 0.286 \\ 
 \hline

\end{tabular}
\caption{Emotional salience extracted from six examples of emotion elicitation-related propaganda techniques.}
\label{table:one}
\end{table}
% \vspace{5pt}

% To gain an initial sense this main design intuition, we applied CrystalFeel and derived the emotion intensity scores for five examples used in \newcite{EMNLP19DaSanMartino} (see Appendix A). This initial analysis suggests that these examples can be potentially characterized with a moderate to high degree of emotion intensities (intensity score $>$ 0.4). 

% In the following sections, we present our emotional salience-based system which differentiates with existing work that mainly leverage neural models such as LSTM \cite{rashkin2017truth} and BERT features \cite{EMNLP19DaSanMartino}. 
%In development dataset, we found that our system incorporating emotion intensity features and BERT features trained with a simple feedforward network, obtained micro-averaged F1-score of 0.570 on overall detection efficacy over 14 propaganda techniques, which is superior to BERT-only based system. Evaluated on gold test data, our system performed relatively well (F1 $>$ 0.5) in detecting loaded language technique, name calling and labeling technique, doubt and flag waving techniques, where most of these are conceptually associated with emotions.

\section{Related Work}

% \textbf{The task setup.} The main task we participated in is the Technique Classification task (task TC). This task aims to classify each given text segment for each of the fourteen propaganda techniques. The input data is a text segment marked with superscripts indicating the start and the end characters that are supposed to be classified. For each text segment, the output should be a classification result that marks the existence of one or more of the fourteen propaganda techniques.

% It is useful to note that the class distribution for most of the techniques is highly imbalanced: 11 of the 14 techniques have less than 10\% occurrences over the total 1,043 text segments (see Appendix B). Some techniques such as ``bandwagon,reductio\_ad\_hitlerum" (0.5\%), ``appeal to authority" (1.3\%), ``thought-terminating cliches" (1.6\%) have less than 2\% occurrence. ``loaded language" has most occurrence (30.7\%), followed by ``name calling,labeling" (17.5\%) and ``repetition" (12.6\%). Most text segments have one corresponding technique, but some may have more than one techniques. For example, text segment ``She's a big fan of torture" from article (id = 738361208, span\_start = 2396, span\_end = 2422 has two gold labels ````exaggeration,minimisation"" and ``name calling,labeling".

\textbf{Propaganda analysis, system and detection.} Computational approach to propaganda detection is relatively a new topic (see \newcite{IJCAI2020DaSanMartino} for a review). \newcite{EMNLP19DaSanMartino} formulate the problem of the detection of specific propaganda techniques which is directly related to this paper. \newcite{barron2019proppy} and \newcite{ACL20DaSanMartino} showed how their Proppy and Prta systems can support users to unmask and analyze propaganda in the news with interactive interfaces.

The closest to our work is the analytic study by \newcite{rashkin2017truth}.  \newcite{rashkin2017truth} compared the linguistic patterns, e.g., psycholinguistic features from LIWC, sentiments, hedging words and intensifying words, across four categories of news: propaganda, trusted news, hoax, or satire. They found interesting linguistic differences in the three “fake” news categories vis-à-vis trusted news, though the predictive experiments showed that LIWC do not improve over the neural model in terms of predictive model performance, probably due to that “some of this lexical information is perhaps redundant to what the model was already learning from the text” \cite{rashkin2017truth}. What \newcite{rashkin2017truth} focused on are word-level or lexical linguistic features. None of the existing work has explored the value of sentence-level sentiment and emotion intensity features in the context of propaganda detection. 

\textbf{Emotion intensity detection and analysis.} Classic sentiment analysis typically provides classification results for discrete sentiment (e.g., positive, negative, neutral) and emotion classification analysis (e.g., happy vs. no happy, sad vs. no sad). Emotion intensity analysis is relatively a new development in the context of predicting the degree or intensity of the underlying emotional valence and dimensions in text messages such as tweets \cite{mohammad2017emotion,mohammad2018semeval}. \newcite{gupta2018crystalfeel} trained CrystalFeel with features derived from parts-of-speech, n-grams, word embedding, multiple existing affective lexicons, and an in-house developed emotion intensity lexicon to predict the degree of the intensity associated with fear, anger, sadness, and joy in the tweets. Its predicted sentiment intensity had arrived a Pearson correlation coefficient ($r$) value of .816 on sentiment intensity with out-of-training sample of human annotations, and of .708, .740, .700 and .720 on emotion intensities in predicting joy, anger, fear and sadness \cite{gupta2018crystalfeel}. %When applied as features in related predictive tasks, these emotion intensity scores are found useful features for training and developing new machine learning methods for understanding the ingredients of happiness \cite{DBLP:conf/aaai/GuptaBY19} and predicting the social popularity of news headlines \cite{DBLP:conf/mm/GuptaY19}. 

\section{Correlation Analysis}
To gain an exploratory understanding on the usefulness of the emotional salience features, we performed bivariate correlation analyses between each of the propaganda ground truth labels for the 1,043 text segments in the development set and the emotion intensity scores derived from CrystalFeel. Table \ref{table:correlation} reports the correlation results. Non-parametric measure of Kendall's $\tau$ was used for the correlation test because the ground truth is a dichotomous variable (1 indicates the propaganda technique is present in the text; 0 indicates otherwise).

\begin{table}[h!]
\centering
 \begin{tabular}{| p{0.32\linewidth} | C{0.1\linewidth} | C{0.1\linewidth}  | C{0.1\linewidth}  | C{0.1\linewidth} | C{0.1\linewidth}  |}
 \hline
 \multirow{2}{*}{\textbf{Propaganda technique}}
      & \multicolumn{5}{c|}{\textbf{Kendall's $\tau$ coefficient}}\\
      \cline{2-6}
%  \hline

  & Valence Intensity & Joy Intensity & Anger Intensity & Fear Intensity & Sadness Intensity \\  
 \hline
 
 ``appeal to authority"  & 0.041 & 0.011 & -0.002 & -0.010 & \textbf{-0.066**} \\
 \hline
  \textbf{``appeal to fear-prejudice"}  & \textbf{-0.064*} & \textbf{-0.074**} & \textbf{0.073**} & \textbf{0.160**} & \textbf{0.057*} \\
 \hline
  ``bandwagon,reductio\_ad\_hitlerum" &  0.042 & 0.028 & 0.001 & 0.030 & \textbf{-0.060*} \\
 \hline
  ``black-and-white fallacy"  & 0.019 & -0.019 & -0.039 & -0.038 & -0.041 \\
 \hline
  ``causal oversimplification"   & -0.004 & -0.035 & 0.046 & 0.022 & -0.012 \\
 \hline
  ``doubt"  & \textbf{-0.079**} & \textbf{-0.118**} & \textbf{0.071**} & 0.045 & -0.023 \\
 \hline
  ``exaggeration,minimisation"  & \textbf{0.051*} & \textbf{0.086**} & -0.012 & -0.014 & 0.010 \\
 \hline
 \textbf{``flag waving"}  & \textbf{0.182**} & \textbf{0.099**} & \textbf{-0.179**} & \textbf{-0.168**} & \textbf{-0.167**} \\
 \hline
 \textbf{``loaded language"}  & \textbf{-0.224**} & \textbf{-0.089**} & \textbf{0.181**} & \textbf{0.140**} & \textbf{0.243**} \\
 \hline
 ``name calling,labeling"  & \textbf{0.066**} & 0.032 & -0.039 & 0.010 & \textbf{-0.062*} \\
 \hline
 ``repetition"  & 0.032 & 0.029 & \textbf{-0.066**} & \textbf{-0.081**} & 0.006 \\
 \hline
 \textbf{``slogans"}  & \textbf{0.089**} & \textbf{0.063*} & \textbf{-0.110**} & \textbf{-0.136**} & \textbf{-0.088**} \\
 \hline
 ``thought-terminating cliches"   & \textbf{0.063*} & \textbf{0.065*} & \textbf{-0.062*} & \textbf{-0.050*} & -0.040 \\
 \hline
 ``whataboutism,straw men,red h."  &  0.015 & -0.009 & 0.011 & -0.005 & \textbf{-0.068**} \\
 \hline

\end{tabular}
\caption{Correlation between the ground truth labels and emotion intensities in the development set \\(** indicates p value $< 0.01$; * indicates p value $< 0.05$)}
\label{table:correlation}
\end{table}

Results indicate interesting patterns: ``loaded language”, ``flag waving", ``slogans", ``appeal to fear-prejudice", and a total of twelve propaganda techniques are significantly correlated with at least one of the emotion intensity scores ($**p < 0.01, *p < 0.05, n=1,043$).

Most notably, ``loaded language" is negatively correlated with valence intensity ($\tau = -0.224$**) and joy intensity ($\tau = -0.089$**), but is positively correlated with anger intensity ($\tau = 0.181$**), fear intensity ($\tau = 0.140$**) and sadness intensity ($\tau = 0.243$**). The ``slogans" technique has a similar correlational pattern.

In contrast, ``flag waving" has the exact opposite pattern, where it is positively correlated with valence intensity ($\tau = 0.182$**) and joy intensity ($\tau = 0.099$**), but is negatively correlated with anger intensity ($\tau = -0.179$**), fear intensity ($\tau = -0.168$**) and sadness intensity ($\tau = -0.167$**). The ``appeal to fear-prejudice" technique has a similar pattern.

Two propaganda techniques, ``black-and-white fallacy" and ``causal oversimplification", are not found to be correlated with any emotion intensity scores. Noted that these techniques also have less occurrences in the dataset (gold labels $< 3\%$; see Appendix A) and are not conceptually associated with emotional connotation or emotional appeal by definition.

The results showed initial support to our main design intuition, which also implies that the emotional saliences based system is likely to be effective in detecting emotions-associated (but not non-emotions-associated) propaganda techniques.

\section{System Overview}

\begin{figure*}[h]
	\centering
	\includegraphics[width=0.68\textwidth]{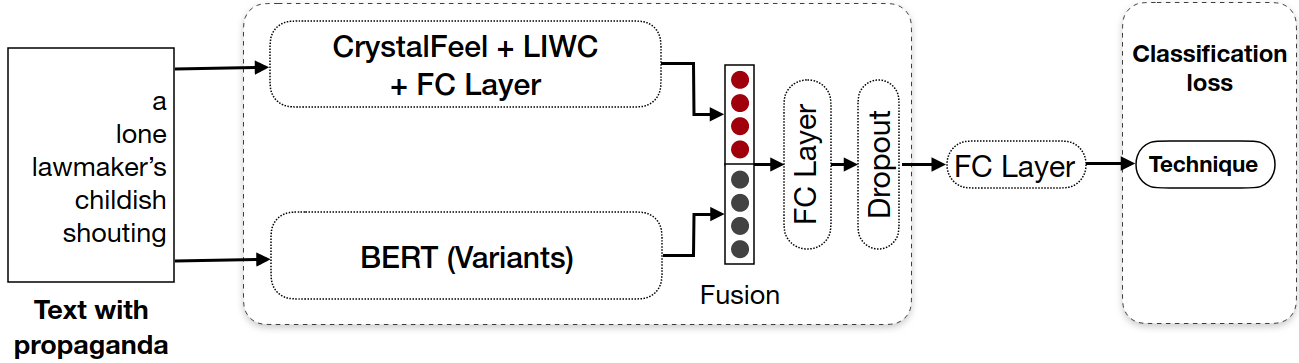}
	\caption[]{Architecture diagram of the proposed model}
	\label{fig:model_fig}
% 	\vspace{-0.5cm}
\end{figure*}

% \subsection{Problem  Formulation}
Following the the exploratory analysis, we proceed to design a predictive system named as ``SocCogCom". Our SocCogCom system is designed to determine the specific propaganda technique used in a given text segment from news articles. The possible techniques are based on a range of fourteen possibilities which are defined in the official SemEval 2020 Task 11 description paper \cite{DaSanMartinoSemeval20task11}. Figure \ref{fig:model_fig} depicts the system architecture.

% \subsection{Model Details}

\textbf{Input Layer:} A training example consists of the span of text that contains a propaganda technique: $x \in \mathbb{R}^{n}$ and a propaganda technique label associated with the text: $y \in$ \{\verb!14 techniques!\}. $x$ is a sequence of words represented in the order of appearance in the vocabulary.

\textbf{Features Extraction:} For every input text segment, our system extracts the following features: 
% \vspace{-3pt}
\begin{tight_enumerate}
\item \textbf{BERT features\footnote{https://storage.googleapis.com/bert\_models/2018\_10\_18/uncased\_L-24\_H-1024\_A16.zip}:} Sentence-level embeddings ($b_f$) \cite{devlin2018bert}. This is a set of pre-trained sentence-level embedding features with a total of 1,024 dimensions.

\item \textbf{CrystalFeel features\footnote{http://www.crystalfeel.socialanalyticsplus.net}:} Sentence-level emotional saliences features \cite{gupta2018crystalfeel} ($c_f$). The extracted features for each text segment include five dimensions of emotion intensity features.
%tValence (real-valued valence intensity in the range of 0 to 1, where 0 means extremely unpleasant and 1 means extremely pleasant), tJoy (joy intensity from 0 to 1), tAnger (anger intensity from 0 to 1), tFear (fear intensity from 0 to 1), tSadness (sadness intensity from 0 to 1), and two dimensions of tEmotionScore (a normalized main emotion prediction result in the range of 1 to 10, where 1 means no emotion and 10 means extremely intense emotion is present) and tSentimentScore (a categorical sentiment score where -2 indicates very negative, -1 negative, 0 neutral or mixed, 1 positive, and 2 very positive).

\item \textbf{LIWC Features:} Word-level psycholinguistic features from the LIWC lexicon\footnote{https://liwc.wpengine.com} \cite{pennebaker2015development} ($l_f$). We obtained 73 extracted features that represent psycholinguistic characteristics of a piece of text that may involve a propaganda technique.
\end{tight_enumerate}

% \begin{enumerate}
%     \item BERT (\hl{cite}) features. ($b_f$)
%     \item Emotional Features from CrystalFeel. ($c_f$)
%     \item Psycholinguistic features from LIWC Lexicon. ($l_f$)
% \end{enumerate}

\vspace{-5pt}
\textbf{Fusion Layer:} CrystalFeel and LIWC features obtained above are first concatenated and a dense layer is applied over the concatenated vector to obtain a feature vector, $h_f$, of dimension $d_h = 50$. This is done in order to project the features extracted from CrystalFeel and LIWC to a similar latent space as that of BERT features. Here, the extracted features, $b_f$ and $h_f$, are simply concatenated to form the representation: $z_f = [b_f; h_f]$ of dimension $d_{in} = 1074$. A dense layer with 256 dimensions is applied over $z_f$. After this, the final representation - $o_f$ is obtained by applying a dropout layer \cite{dropout2014} with dropout rate of 0.5.

\textbf{Output Layer:} The system employs a fully-connected layer with softmax activation where the fused representation $o_f$ is fed.

\textbf{Loss function:} The categorical cross-entropy is used to calculate the loss. We minimize the loss with an optimizer. The function that is optimized is as follows:
\begin{equation}
  \label{eq:categoricalcrossentropy}
  E_{\mbox{\tiny crossentropy}} = -\sum_{n=1}^N \sum_{k=1}^c
  y_k^n \log \hat{y}_k^n
\end{equation}
where $N$ is the total number of samples and $c$ is the number of classes (in our case it is 14). $y_k^n$ is the actual label of the $k^{th}$ class of the $n^{th}$ sample and $\hat{y}_k^n$ is the prediction corresponding to the $k^{th}$ class of the $n^{th}$ sample.

% \begin{equation}
%   \label{eq:categoricalcrossentropy}
%   E_{\mbox{\tiny crossentropy}} = -\sum_{n=1}^N \sum_{k=1}^c
%   y_k^n \log \hat{y}_k^n
% \end{equation}
% where $N$ is the total number of samples and $c$ is the number of classes (in our case it is 14). $y_k^n$ is the actual label of the $k^{th}$ class of the $n^{th}$ sample and $\hat{y}_k^n$ is the prediction corresponding to the $k^{th}$ class of the $n^{th}$ sample.

\section{Features Experiments and Results}
% In this section, we will discuss about the experimental setup for the technique classification task.
% \subsection{Dataset}
% We use the official dataset provided by the organizers of SemEval 2020 Task 11 (\hl{cite}). And more specifically, only the dataset provided for the second sub-task: technique classification. 
% In table \ref{tab:data-distri} we show the distribution of data in this dataset.

% \begin{table}[]
% \centering
% \begin{tabular}{|l|c|c|c|}
% \hline
%                  & \multicolumn{3}{c|}{\textbf{Dataset}}      \\ \hline
% \textbf{Technique} & \textbf{Train} & \textbf{Dev} & \textbf{Test} \\ \hline
% Red Herring   &            &           &            \\ \hline
% Straw Man    &            &           &            \\ \hline
% Whataboutism    &            &           &          \\ \hline
% Causal Oversimplification &           &          &           \\ \hline
% Obfuscation &           &          &           \\ \hline
% Appeal to authority &           &          &           \\ \hline
% Black-and-white Fallacy &           &          &           \\ \hline
% Name calling &           &          &           \\ \hline
% Loaded Language &           &          &           \\ \hline
% Exaggeration or Min. &           &          &           \\ \hline
% \end{tabular}
% \caption{Distribution of technique class labels in Train, Development and Test sets in the dataset.}
% \label{tab:data-distri}
% \end{table}

% \subsection{Data pre-processing}
For data pre-processing, we used Keras Tokenizer to split the text into word tokens. The sentences are cleaned to remove unwanted characters and double spaces are replaced with single space.

% \subsection{Emotional Features}
% We extract emotional features of the input text from CrystalFeel. The features are \verb!tAnger!, \verb!tFear!, \verb!tJoy!, \verb!tSadness!, \verb!tValence!, \verb!tEmotionScore! and \verb!tSentimentScore!.

% \subsection{Psycholinguistics Features}
% We use Linguistic Inquiry and Word Count (LIWC) lexicon to extract Psycholinguistics features of a span of text. We use 73 extracted features that represents psycholinguistic characteristics of a piece of text that has a propaganda technique.

% \subsection{Model training}
We conducted the features experiments using the standard training and development datasets provided in the official TC task, based on the system set up described in Section 3. Hyper-parameters are tuned using a held out validation data: $10\%$ of the training data. To optimize the parameters, we use Adam optimizer ~\cite{AdamKingmaB14} with an initial learning rate of $1e^{-4}$. The experiments results are presented in Table \ref{table:ablation}.

%The hyper-parameter are $f_l=128$, $M = 3$. And for each of the 3 filter size the window size ($h$) is 2, 3 and 4. We fix the tweet length $n$ to 30 and target length $m$ to 6. The number of hidden units in task-specific layers $fc_{[1/2/3]}$ is 300. We initialize the word vectors with the $300$-dimensional pre-trained word2vec embeddings~\cite{mikolov2013distributed} which are optimized during training. Following the previous works, we train different models for different targets but with the same hyperparameters. And the final result is the concatenation of predicted result of these models.

% \subsection{Evaluation metrics}
% Technique Classification is a multi-class classification task. The official evaluation metric is the Micro-averaged F1 score ($F1_\mu$) of all the technique classes (see Appendix C).

% \subsection{Feature Experiments Results}

% We used the official TC training and development datasets to evaluate the effectiveness of the features, based on the system setup described in Section 3.

\begin{table}[h!]
\centering
 \begin{tabular}{| l | c |}
 \hline
 \textbf{Model + Features} & \textbf{Micro-averaged F1 score} \\
 \hline
 Logistic Regression & 0.2520\\
 \hline
 BERT Only & 0.5485  \\ 
 \hline
 CrystalFeel Only & 0.5234\\
 \hline
 \textbf{BERT + CrystalFeel} & \textbf{0.5701}  \\
 \hline
 BERT + CrystalFeel + LIWC & 0.5626  \\
 \hline
 AlBERT + CrystalFeel & 0.5588 \\
 \hline
 \textbf{BERT + CrystalFeel + Context} & \textbf{0.5824} \\
 \hline
\end{tabular}
\caption{Feature experiments results on development set.}
\label{table:ablation}
\end{table}

First, we evaluated the effects of using BERT features and emotional salience features from CrystalFeel outputs alone. BERT only obtained micro-averaged F1 score of 0.5484, showing strong performance in comparison to a simple baseline using logistic regression. CrystalFeel features achieved 0.5234, which shows fair performance given this is a low-dimensional features set. When combined, BERT and CrystalFeel features achieved better results, with micro-averaged F1 score of 0.5701 than the individual settings.

We also assessed classic word-level psycholinguistics features based on LIWC lexicons. The BERT + LIWC condition didn't converge, as the loss didn't decrease and was fluctuating a lot. Adding LIWC onto the hybrid BERT + CrystalFeel, i.e., the BERT + CrystalFeel + LIWC condition, obtained micro-averaged F1 score of 0.5626, indicating that additional word-level psycholinguistics features do not appear to improve over the BERT + CrystalFeel condition. We tested AlBERT + CrystalFeel too, and they did not match the results obtained from BERT + CrystalFeel condition.

% \vspace{-5pt}
Based on the experiment results, we used the best-performing hybrid features sets (BERT + CrystalFeel) for our system results submission for the gold test set. 

After we submitted our results, we experimented a new condition where context features were added to the BERT + CrystalFeel condition. For context, we extracted features using 3 words before and after the target text segment. The results showed improvement (micro-averaged F1 score = 0.5824).

\section{Results on Gold Test Set}

Overall, on gold test set, the results released from the task organizers suggested that our system achieved micro-averaged F1 score of 0.558 across the fourteen propaganda techniques. Table \ref{table:detailed} shows the detailed results showing F1 scores of our system for each propaganda technique.

\begin{table}[h!]
\centering
 \begin{tabular}{| p{0.37\linewidth} | C{0.1\linewidth} | C{0.13\linewidth}  | C{0.1\linewidth}  | C{0.13\linewidth} |}
 \hline
 \multirow{2}{*}{\textbf{Propaganda techniques}}
      & \multicolumn{2}{c|}{\textbf{F1 (development set)}}
      & \multicolumn{2}{c|}{\textbf{F1 (gold test set)}}\\  
      \cline{2-5}
%  \hline

  & Baseline & SocCogCom & Baseline & SocCogCom \\  
 \hline
 
 “appeal to authority”  & 0.000 & 0.000 & 0.000 & 0.286 \\
 \hline
  ``appeal to fear-prejudice"  & 0.094 & 0.329 & 0.037 & 0.316\\
 \hline
  ``bandwagon,reductio\_ad\_hitlerum"   & 0.000 & 0.571 & 0.000 & 0.098\\
 \hline
  ``black-and-white fallacy"   & 0.000 & 0.214 & 0.000 & 0.265 \\
 \hline
  ``causal oversimplification"   & 0.072 & 0.286 & 0.116 & 0.063 \\
 \hline
  \textbf{``doubt"}  & 0.192 & \textbf{0.540} & 0.291 &\textbf{ 0.604} \\
 \hline
  ``exaggeration,minimisation"  & 0.117 & 0.457 & 0.144 & 0.349 \\
 \hline
 \textbf{``flag waving"}  & 0.083 & \textbf{0.771} & 0.062 & \textbf{0.543} \\
 \hline
 \textbf{``loaded language"}  & 0.406 & \textbf{0.706} & 0.465 & \textbf{0.722}\\
 \hline
 \textbf{``name calling,labeling"}  & 0.000 & \textbf{0.644} & 0.000 & \textbf{0.673}\\
 \hline
 ``repetition"  & 0.385 & 0.318 & 0.193 & 0.189\\
 \hline
 ``slogans"  & 0.000 & 0.302 & 0.000 & 0.409\\
 \hline
 ``thought-terminating cliches"   & 0.000 & 0.129 & 0.000 & 0.235\\
 \hline
 ``whataboutism,straw men,red herring"  &  0.000 & 0.064 & 0.000 & 0.100\\
 \hline
 \textbf{Micro-averaged F1}   & 0.265 & \textbf{0.570} & 0.252 &\textbf{ 0.558}\\
 \hline
\end{tabular}
\caption{Predictive results of our system for each propaganda techniques.}
\label{table:detailed}
\end{table}

The results suggested that using relatively parsimonious features, BERT and CrystalFeel emotional salience features, our system performed reasonably well (F1 score $>$ 0.5) in detecting ``loaded language" (F1 = 0.772), ``name calling and labeling" (0.673), ``doubt" (0.604) and ``flag waving" (0.543). Meanwhile, our system struggled in detecting non-emotion associated techniques (which also happen to have imbalanced distributions), such as ``causal oversimplification" (F1 = 0.063), ``bandwagon,reductio\_ad\_hitlerum" (F1 = 0.098), and ``whataboutism, straw men, red herring" (F1 = 0.100). The results also support with our design intuition that the sentiment and emotion intensities features help to detect propaganda techniques which are manifested in their emotional salience in the text segment.

\section{Conclusion}
\label{sec:conclusion}
Propaganda is primarily information that is used to advance an agenda through influence techniques. Our work is motivated to explore the value of emotional salience features in predicting emotion-related propaganda techniques. In our experiments, we found that emotional salience features using CrystalFeel emotion intensity scores can improve over BERT only features, when a simple feedforward neural network is used in both experiment settings. Results and analysis on gold test dataset show that our approach performed reasonably well (F1 $>$ 0.5) in detecting ``loaded language", ``name calling and labeling", ``doubt" and ``flag waving" techniques. As these are also most frequently used techniques, our system has a potential value to facilitate publishers and general public to be alerted with these common techniques. The system scripts are released at \url{https://github.com/gangeshwark/PropagandaNews}.

% We have released our system and extracted features - \url{https://github.com/gangeshwark/PropagandaNews} - for potential applications and further research.

\section*{Acknowledgements}

This work is supported by Agency for Science, Technology and Research (A*STAR) under its A*ccelerate Gap Fund (ETPL/18-GAP051-R20A).

\bibliographystyle{coling}
\bibliography{semeval2020}

\newpage

\section*{Appendix A: Propaganda techniques, definitions and the gold labels distributions}

The task TC aims to classify each given text segment for each of the fourteen propaganda techniques. The input data is a text segment marked with superscripts indicating the start and the end characters that are supposed to be classified. For each text segment, the output should be a classification result that marks the existence of one or more of the fourteen propaganda techniques.

Most text segments have one corresponding technique, but some may have more than one techniques. For example, text segment ``She's a big fan of torture" from article (id = 738361208, span\_start = 2396, span\_end = 2422) has two gold labels ``exaggeration,minimisation" and ``name calling,labeling".

It is useful to note that the class distribution for most of the techniques is highly imbalanced: 11 of the 14 techniques have less than 10\% occurrence over the total 1,043 text segments (see table below for details). Some techniques such as ``bandwagon,reductio\_ad\_hitlerum" (0.5\%), ``appeal to authority" (1.3\%), ``thought-terminating cliches" (1.6\%) have less than 2\% occurrence. ``loaded language" has most occurrence (30.7\%), followed by ``name calling,labeling" (17.5\%) and ``repetition" (12.6\%).

% \begin{longtable}[]
% \centering
%  \begin{tabular}{| p{0.3\linewidth} | p{0.1\linewidth} | p{0.1\linewidth}  | p{0.1\linewidth}  | p{0.1\linewidth}  | p{0.1\linewidth}  |}
\begin{longtable}{| C{0.2\linewidth} | p{0.45\linewidth} | C{0.1\linewidth}  | C{0.1\linewidth} |}
 \hline
%  \multirow{2}{*}{\textbf{Propaganda techniques}}
    %   & \multicolumn{5}{c|}{\textbf{Detected Emotion Intensity Scores (Gupta and Yang, 2018)}}\\  
    %   \cline{2-6}
%  \hline

  Propaganda techniques & Definitions \cite{EMNLP19DaSanMartino} & Gold labels (count) & Gold labels (\%)   \\  
  \hline
 
``appeal to authority" & Stating that a claim is true simply because a valid authority/expert on the issue supports it, without any other supporting evidence (Goodwin, 2011) & 14 & 1.3\%  \\ 
 \hline
``appeal to fear-prejudice"*  &  Seeking to build support for an idea by instilling anxiety and/or panic in the population towards an alternative, possibly based on preconceived judgments & 44 & 4.2\%  \\ 
 \hline
 
``bandwagon, reductio\_ad\_hitlerum" & \textbf{Bandwagon}: Attempting to persuade the target audience to join in and take the course of action because “everyone else is taking the same action” (Hobbs and Mcgee, 2008). \textbf{Reductio\_ad\_hitlerum}: Persuading an audience to disapprove an action or idea by suggesting that the idea is popular with groups hated in contempt by the target audience. It can refer to any person or concept with a negative connotation (Teninbaum, 2009) & 5 & 0.5\%  \\ 
 \hline
 
``black-and-white fallacy" & Presenting two alternative options as the only possibilities, when in fact more possibilities exist (Torok, 2015). As an extreme case, telling the audience exactly what actions to take, eliminating any other possible choice (dictatorship) & 22 & 2.1\% \\
 \hline
 
``causal oversimplification" & Assuming one cause when there are multiple causes behind an issue. & 18 & 1.7\% \\
\hline
``doubt" & Questioning the credibility of someone or something & 66 & 6.3\% \\
\hline
``exaggeration, minimisation"* & Either representing something in an excessive manner: making things larger, better, worse or making something seem less important or smaller than it actually is \cite{jowett2018propaganda} & 68 & 6.5\% \\
\hline
``flag waving"* & Playing on strong national feeling (or with respect to a group, e.g., race, gender, political preference) to justify or promote an action or idea (Hobbs and Mcgee, 2008) & 86 & 8.2\% \\
\hline
``loaded language"* & Using words/phrases with strong emotional implications (positive or negative) to influence an audience (Weston, 2018, p. 6) & 320 & 30.7\% \\ 
\hline
``name calling,labeling"* & Labeling the object of the propaganda campaign as either something the target audience fears, hates, finds undesirable or otherwise loves or praises (Miller, 1939) & 183 & 17.5\%  \\
\hline
``repetition" & Repeating the same message over and over again, so that the audience will eventually accept it (Torok, 2015; Miller, 1939) & 131 & 12.6\% \\
\hline
``slogans"* & A brief and striking phrase that may include labeling and stereotyping. Slogans tend to act as emotional appeals (Dan, 2015) & 40 & 3.8\% \\
\hline
``thought-terminating cliches" & Words or phrases that discourage critical thought and meaningful discussion about a given topic. They are typically short, generic sentences that offer seemingly simple answers to complex questions or that distract attention away from other lines of thought (Hunter, 2015, p. 78). &17 & 1.6\% \\
\hline
``whataboutism, straw men, red herring" & \textbf{Whataboutism}: Discredit an opponent's position by charging them with hypocrisy without directly disproving their argument (Richter, 2017). \textbf{Straw Men}: When an opponent's proposition is substituted with a similar one which is then refuted in place of the original (Walton, 1996). \textbf{Red Herring}: Introducing irrelevant material to the issue being discussed, so that everyones attention is diverted away from the points made (Weston, 2018, p. 78) & 29 & 2.8\% \\
\hline
% \label{table:intro}
\caption{Propaganda techniques, definitions and the gold labels distributions in the development set (total n=1,043 text segments)}
\end{longtable}
% \end{longtable}
* These six techniques are associated with emotions by their respective definitions.

\end{document}